\documentclass{article}
\usepackage{microtype}
\usepackage{graphicx}
\usepackage{subcaption}
\usepackage{booktabs} % for professional tables

\usepackage{hyperref}

\usepackage[preprint]{icml2026}

\usepackage{amsmath}
\usepackage{amssymb}
\usepackage{mathtools}
\usepackage{amsthm}

\usepackage[capitalize,noabbrev]{cleveref}

\theoremstyle{plain}

\theoremstyle{definition}

\theoremstyle{remark}

\usepackage[textsize=tiny]{todonotes}

\usepackage{wrapfig}
\usepackage{subcaption}
\usepackage{graphicx}
\usepackage{float}
\usepackage[most]{tcolorbox}
\usepackage{xcolor}
\usepackage[capitalize,noabbrev]{cleveref}
\usepackage{enumitem}
\usepackage{lipsum}

\newcommand{\tocite}[1]{{\color{red} CITE}}

\newcommand{\synthetic}{\textsc{Sieve-Gen}}
\newcommand{\method}{\textsc{Sieve}}

\icmltitlerunning{Sample-Efficient Parametric Learning from Natural Language}

\begin{document}

\twocolumn[
  \icmltitle{\method{}: Sample-Efficient Parametric Learning from Natural Language}



  \icmlsetsymbol{equal}{*}

  \begin{icmlauthorlist}
    \icmlauthor{Parth Asawa}{berkeley}
    \icmlauthor{Alexandros G. Dimakis}{berkeley,bespoke}
    \icmlauthor{Matei Zaharia}{berkeley}
  \end{icmlauthorlist}
    
  \icmlaffiliation{berkeley}{University of California, Berkeley, CA, USA}
  \icmlaffiliation{bespoke}{Bespoke Labs}
    
    % Corresponding author info (only appears in accepted version)
  \icmlcorrespondingauthor{Parth Asawa}{pgasawa@berkeley.edu}

  \icmlkeywords{Large Language Models, Steering, Alignment, Reinforcement Learning, Black-Box Optimization}

  \vskip 0.3in
]



\printAffiliationsAndNotice{}  % no special notice (required even if empty)

\begin{abstract}
Natural language context—such as instructions, knowledge, or feedback—contains rich signal for adapting language models. While in-context learning provides adaptation via the prompt, parametric learning persists into model weights and can improve performance further, though is data hungry and heavily relies on either high-quality traces or automated verifiers. We propose \method{}, a method for sample-efficient parametric learning from natural language context that requires as few as three query examples. \method{} uses a novel synthetic data generation pipeline, \synthetic{}, that leverages the insight that context is \textit{decomposable}. Decomposing context allows us to generate higher quality rollouts by pairing synthetic queries with only the applicable context rather than the entirety, then using context distillation to internalize context into the model. We evaluate in reasoning settings where context is necessary, including custom domains and the RuleArena and Machine Translation from One Book tasks. Our results show that \method{} outperforms prior context distillation methods using just three query examples, demonstrating how to achieve sample-efficient parametric learning from natural language.
\end{abstract}

\section{Introduction}
\label{sec:introduction}

Language models today rely heavily on in-context learning (ICL) to adapt to new tasks: users provide examples, instructions, feedback, or domain knowledge directly in prompts to guide model behavior \citep{brown2020languagemodelsfewshotlearners}. This approach has become ubiquitous—from developers writing extensive system prompts with coding standards and API documentation, to users providing personal preferences and writing styles, to domain experts supplying specialized knowledge for medical diagnosis or legal reasoning \citep{xu2024samdkifscalableadaptablemedical, liu2025surveypersonalizedlargelanguage, sahoo2025systematicsurveypromptengineering}. However, ICL has fundamental limitations: it cannot leverage the benefits of parametric learning, such as eliminating context window constraints, enabling persistent improvements that survive across sessions, and improving performance through additional training compute.

Due to these limitations, a line of work has emerged exploring parametric learning—directly internalizing context into model weights through training. Context distillation approaches \citep{snell2022learningdistillingcontext, bhargava2024promptbaking, upadhayayaya2025efficientllmcontextdistillation} ``bake" instructions and examples into weights by training a student model to imitate a teacher with access to context. Methods for training with natural language feedback \citep{scheurer2022traininglanguagemodelslanguage, chen2024improvingcodegenerationtraining} show that textual corrections can improve model outputs. However, these parametric methods face a critical bottleneck: they are data-hungry, typically requiring many examples of queries or expensive expert-generated traces and automated verifiers. This creates a wide gap: ICL works with minimal examples but cannot access parametric benefits, while parametric methods offer these benefits but demand abundant data. \textbf{Can we achieve the advantages of parametric learning with the sample efficiency of in-context learning?}

We demonstrate that the answer is yes. With \textbf{as few as three examples of task queries}, models can internalize natural language context requiring multi-step reasoning, outperforming prior context distillation methods and even matching or exceeding ICL performance without needing context at inference time.

To create a sample-efficient method, we seek to leverage synthetic queries and rollouts that are created from a natural language context ($\mathcal{C}$) and very few examples of queries. Our key insight is that \textbf{natural language context is decomposable}. Natural language context often consists of independent context units ($u$) where only a subset applies to any given query. For example, a list of rules can be broken into individual parts, with each task requiring only a few; a grammar specification decomposes into individual constraints, with each translation needing only relevant ones. The quality of context distillation training is heavily linked to the quality of the rollouts with context ($r$). Decomposing context allows us to filter and only include the context units that are applicable to a particular query ($c_a \subseteq \mathcal{C}$), leading to higher quality rollouts for training than prior methods that indiscriminately provide all context for all queries.

Using this, we propose \method{}, a method that achieves sample-efficient parametric learning from natural language context. The heart of our approach is \textbf{\synthetic{}}, a novel synthetic data generation pipeline that requires natural language context and just a few examples of queries. \synthetic{} breaks context into context units, leverages a base language model to create diverse sets of context units for generating synthetic queries, then filters which units actually apply to each query. This produces training data where queries are paired with \textit{only} their applicable context, yielding higher-quality rollouts. We then apply standard context distillation techniques \citep{snell2022learningdistillingcontext}, distilling the model's behavior when conditioned on applicable context into weights that can perform the same reasoning without context. Avoiding prior limitations, \method{} enables parametric learning from natural language both without access to many examples of queries over context nor high quality traces or verifiers.

We evaluate \method{} on domains requiring reasoning over context (not just factual recall): our Retail domain tests compositional rule application over 30 discount rules, RuleArena (NBA) \citep{zhou2025rulearenabenchmarkruleguidedreasoning} evaluates complex sports regulation reasoning, and MTOB \citep{tanzer2024benchmarklearningtranslatenew} requires translating extremely low-resource languages from grammar specifications. Unlike prior work on parametric internalization that focuses on memorizing facts \citep{eyuboglu2025cartridgeslightweightgeneralpurposelong, lin2025learningfactsscaleactive, yang2024syntheticcontinuedpretraining}, these tasks require models to reason over internalized context at inference time. 

\textbf{Our contributions are:}

\begin{enumerate}
    \item We demonstrate that sample-efficient parametric learning from natural language context is achievable with \textit{as few as three task examples}, bridging the gap between ICL's sample efficiency and parametric learning's advantages.
    
    \item We introduce \synthetic{}, a novel synthetic data generation method that exploits context decomposability to create diverse, high-quality training data by pairing queries with only their applicable context.
    
    \item We show empirically that models trained with \method{} outperform prior context distillation methods and can match or exceed in-context learning performance without context at inference time, across multiple reasoning domains and model families.
\end{enumerate}

These findings establish that parametric learning can be practical for incorporating natural language context, enabling persistent improvements from minimal input.
\section{Related Work}
\label{sec:related_work}

\textbf{Context Distillation.} Work in context distillation, also known as prompt baking or prompt distillation, seeks to internalize information from prompts directly into model weights \citep{bhargava2024promptbaking, upadhayayaya2025efficientllmcontextdistillation, snell2022learningdistillingcontext, shenfeld2026selfdistillationenablescontinuallearning}. These methods typically employ a teacher-student framework, where a student model is trained to mimic the output distribution of a teacher model that has access to additional context. While sharing the goal of parametric internalization, all of these works assume access to a distribution of queries or expert demonstrations (traces) from which to sample rollouts. Our work differs by focusing on \textbf{sample-efficient learning from only minimal query examples}.

\textbf{Training with Natural Language Feedback.} Several approaches integrate natural language context during training \citep{scheurer2022traininglanguagemodelslanguage, chen2024improvingcodegenerationtraining, hübotter2026reinforcementlearningselfdistillation}. Rather than relying on scalar preference scores, these methods use rich textual feedback to generate refined outputs for training. For instance, \citet{scheurer2022traininglanguagemodelslanguage} uses human feedback to generate multiple refinements and train on the highest-quality version. While we also adopt this pattern of using natural language context to create training examples, our contribution focuses on achieving this under extreme data constraints \textbf{without requiring expert-curated traces, verifiers, or task-specific datasets}.

\textbf{Synthetic Data for Parametric Knowledge Injection.}  
Recent work explores using synthetic data to inject knowledge into model parameters. Cartridges and related knowledge injection methods \citep{eyuboglu2025cartridgeslightweightgeneralpurposelong, kujanpää2025efficientknowledgeinjectionllms} generate synthetic conversations from long-context document corpora and apply context distillation or lightweight adapters to compress knowledge into KV caches or LoRAs. Active Reading and Synthetic Continued Pretraining \citep{lin2025learningfactsscaleactive, yang2024syntheticcontinuedpretraining} train models to acquire factual knowledge via self-generated synthetic data during pretraining. These approaches primarily target fact recall and memorization, evaluating whether models can retrieve stored information from parameters. In contrast, our work studies internalizing natural language context that requires multi-step reasoning and selective applicability at inference time. Our method and benchmarks therefore emphasize \textbf{reasoning over context}, rather than simply recalling stored facts.

\begin{figure*}[t!]
  \vskip 0.15in
  \begin{center}
  \centerline{\includegraphics[width=0.9\linewidth]{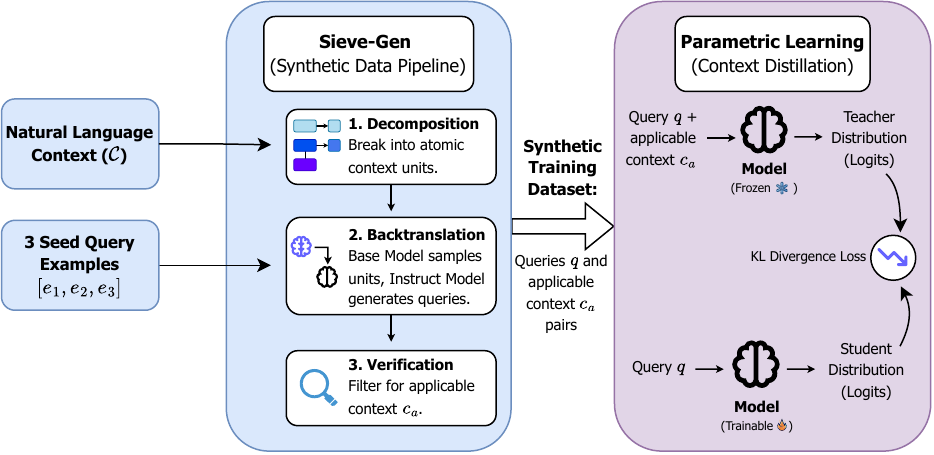}}
  \caption{\textbf{\method{} system overview.} Given a natural language context corpus and as few as 3 seed query examples, \synthetic{} generates synthetic training data composed of (query, applicable context) pairs. These pairs are used for context distillation, where a student model learns to match a teacher's distribution conditioned on applicable context, internalizing the knowledge into weights for inference without context.}
  \label{fig:inscribe}
  \end{center}
  \vskip -0.2in
\end{figure*}
\section{Methods}
\label{sec:methods}

We seek to design a method that integrates natural language context into model weights in a sample-efficient way. Unlike prior work that focuses primarily on memorization or factual recall, our approach targets the internalization of knowledge that requires complex reasoning at inference time. Importantly, our method is \textbf{domain-agnostic}—we use no task-specific prompt engineering, applying the same approach across all domains we evaluate (Appendix \ref{app:synthetic_data_prompts}). Further gains could likely be achieved per-domain by doing so.

Context distillation methods typically make use of many example queries and sample rollouts with the entire context to update the policy. Synthetic data provides an avenue around this, though high quality synthetic data is challenging. As motivated in \cref{sec:introduction}, \method{} leverages a key insight: when integrating natural language context into weights, \textbf{not all context applies to every query}. Simultaneously, natural language context is often compositional; multiple portions may apply to any given query. Decomposing the context we wish to distill into context units enables us to precisely filter context into only what is applicable for particular queries. This filtering leads to higher quality rollouts which in turn leads to higher quality context distillation.

\textbf{Overview.} Leveraging these insights, our approach consists of four steps (\cref{fig:inscribe}):

\begin{enumerate}
    \item Obtain a set of natural language context $\mathcal{C}$ (from humans, automated systems, other LLMs, etc.) relevant to a task or domain, and example queries $[e_1, e_2, e_3]$ that serve as examples to a generator.
    \item Generate synthetic tuples of (query, applicable context units) pairs $(q, c_a)$ where $c_a \subseteq \mathcal{C}$ using \synthetic{}.
    \item Sample a response $r$ from the model conditioned on the query and only the applicable context: $(q, c_a) \rightarrow r$.
    \item Train the model via context distillation to produce $r$ given only $q$, internalizing the context into weights.
\end{enumerate}

\subsection{\synthetic{}: Synthetic Data Generation}

The synthetic data generation process we propose, \synthetic{}, is crucial to the sample efficiency of our method. Unlike typical synthetic data generation that simply samples model outputs for given queries, \synthetic{} needs to both generate diverse queries \textit{and} pair each query with precisely the subset of context units applicable to it, exploiting the decomposability of natural language context.

\synthetic{} operates in three phases: decomposition, backtranslation, and verification, and runs completely offline.

\textbf{Decomposition}. We first prompt an instruction-tuned model to decompose the context corpus $\mathcal{C}$ into atomic context units $\{u_1, u_2, \ldots, u_n\}$ that can be independently evaluated for applicability. Each context unit is intended to be a self-contained piece of knowledge or instruction. This decomposition is performed by the model to ensure context units are semantically coherent (domain-agnostic prompt provided in Appendix \cref{app:synthetic_data_prompts}). Decomposition might turn a list of rules or instructions into individual ones, grammar specifications into individual constraints, etc. We provide example context units for each of the domains we evaluate on in Appendix \ref{app:domain_discussion}.

\textbf{Backtranslation}. To generate diverse synthetic queries, we perform backtranslation from context to queries:

\begin{enumerate}
    \item A base language model (trained only for next-token prediction) samples a subset of context units to serve as a seed: $c_{\text{seed}} \subseteq \{u_1, \ldots, u_n\}$
    \item Given the set of seed context units $c_{\text{seed}}$ and example queries $e_1, e_2, e_3$, an instruction-tuned model generates a synthetic query $q$ for which the seed context units would apply.
\end{enumerate}

The use of a base model for seed selection is critical for diversity. We observed that instruction-tuned models consistently select virtually the exact same subsets of context, leading to narrow coverage of the context space (similar to results shown by \citet{zhu2025bareleveragingbaselanguage}). Following BARE \citep{zhu2025bareleveragingbaselanguage}, we found that base models produce substantially more diverse seeds, resulting in broader internalization of the full context corpus. We utilize a model rather than randomly sampling from the set of context units to ensure coherence in the seed context $c_{\text{seed}}$ from which a query is generated.

\textbf{Verification}. After generating query $q$ from seed context $c_{\text{seed}}$, we verify which context actually applies to this query. 

The model iterates through all context units $\{u_1, \ldots, u_n\}$ and outputs a binary decision for each, producing the verified applicable context $c_a \subseteq \mathcal{C}$. The model is prompted to assess whether each context unit is
necessary to answer the query (see Appendix \ref{app:synthetic_data_prompts}). Along with the rest of \synthetic{}, this verification step is done entirely offline and motivated by the fact that the synthetic query may require additional context beyond the seed, or conversely, not all seed context may actually be necessary.

The resulting tuple $(q, c_a)$ is then used to generate training data as in step 3 of our overview.

\textbf{Long Context Extension.} While we focus our method on natural language context that fits in context for a model to learn from, for settings where context exceeds the model's context window (e.g., MTOB, which uses ~50K tokens for grammar books), we extend \synthetic{} to handle long contexts. We chunk the context corpus by token count (default: 8192 tokens per chunk) with 512-token overlaps to guard against splitting related content. Each chunk is then processed through the decomposition phase independently into a list of context units.

For the verification phase with long contexts, the compute required to generate data by verifying every single point of context individually can become quite large. Instead, we can batch context units together rather than evaluating them individually. Here instead of binary yes/no outputs, the model identifies which specific units from each batch are applicable to the query, reducing computational cost.

\begin{algorithm}[t]
\caption{\synthetic{}: Synthetic Data Generation}
\label{alg:sievegen}
\begin{algorithmic}[1]
\REQUIRE Context $\mathcal{C}$, base model $M_{\text{base}}$, instruction model $M_{\text{inst}}$, num. examples $N$, query examples $[e_1, e_2, e_3]$
\ENSURE Synthetic dataset $\mathcal{D} = \{(q_i, c_{a,i}, r_i)\}_{i=1}^N$
\STATE $\{u_1, \ldots, u_n\} \gets \text{Decompose}(\mathcal{C}, M_{\text{inst}})$
\FOR{$i = 1$ to $N$}
    \STATE $c_{\text{seed}} \gets \text{SampleSubset}(\{u_1, \ldots, u_n\}, M_{\text{base}})$
    \STATE $q_i \gets \text{GenerateQuery}(c_{\text{seed}}, M_{\text{inst}}, [e_1,e_2,e_3])$
    \STATE $c_{a,i} \gets \text{Verify}(q_i, \{u_1, \ldots, u_n\}, M_{\text{inst}})$
    \STATE $r_i \gets M_{\text{inst}}([q_i, c_{a,i}])$
\ENDFOR
\STATE \textbf{return} $\mathcal{D}$
\end{algorithmic}
\end{algorithm}

\subsection{Context Distillation Objective.} 

We base our context distillation objective off of the work of \citet{snell2022learningdistillingcontext}.

Given an input query $q$ and applicable natural language context $c_a \subseteq \mathcal{C}$, the original model $M_\theta$ produces a response conditioned on both $q$ and $c_a$. Rather than training on the model's discrete output, we distill the model's conditional distribution into a student model via soft targets, while removing context from the student input.

Concretely, for each pair of synthetic query ($q$) and applicable context ($c_a$) we generate, we construct a teacher input
\begin{equation}
    c = [q; c_a],
\end{equation}
and obtain the teacher distribution
\begin{equation}
    p_T(y \mid q, c_a) = M_\theta(c).
\end{equation}

We retain the top-$K$ logits from $p_T$ to form a truncated soft target distribution $\tilde{p}_T$, which preserves fine-grained preference information while remaining computationally efficient. We set $k=100$ for our experiments following \citet{snell2022learningdistillingcontext}.

The student model $M_\phi$ is trained on the same query $x$ \emph{without} access to any context $c_a$, and is optimized to match the teacher distribution:
\begin{equation}
    \mathcal{L}_{\text{CD}} = \mathrm{KL}\big( \tilde{p}_T(y \mid q, c_a) \;\|\; M_\phi(y \mid q) \big).
\end{equation}
The model's parameters are optimized over this objective. We experimented with alternative options such as LoRA training though saw worse performance.

\textbf{Models.} We focus our experiments on the Qwen3 8B family of models \citep{qwen3technicalreport}. We provide additional ablations with other model families such as Llama 3.1 8B ~\citep{llama3} and Rnj 1 8B ~\citep{rnj1_base, rnj1_instruct} in \cref{sec:model_families}. We sample rollouts for distillation from the same model we train in all experiments.
\section{Evaluation}
\label{sec:results}

\begin{figure*}[t!]
  \vskip -0.1in
  \begin{center}
  \centerline{\includegraphics[width=\linewidth]{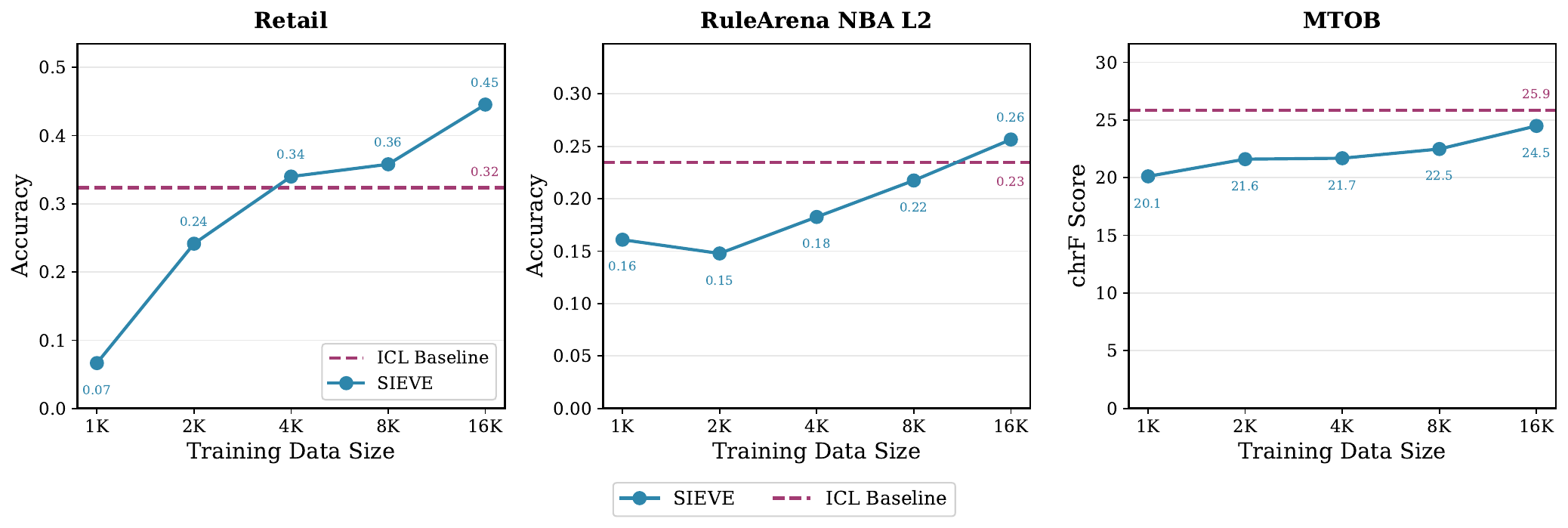}}
  \caption{\textbf{\method{} improves with scale while real data input is constant}. Across various domains, \method{} improves as we scale the amount of data we generate with \synthetic{} (using the same fixed three example queries as inputs), approximately matching or exceeding ICL baseline performance when evaluated without any context. All domains use the Qwen3-8B model family with thinking disabled.}
  \label{fig:scaling_data}
  \end{center}
  \vskip -0.25in
\end{figure*}

We show that with \method{}, natural language context can be efficiently internalized into model weights and reach performance comparable to or surpassing ICL baselines.

\subsection{Domains}
\label{sec:domains}

Because existing benchmarks rarely study adapting from natural language context, we both construct a controlled synthetic domain and modify other benchmarks to fit our setting. We evaluate \method{} on domains requiring reasoning over context (not just factual recall) that are verifiable. We note that we expect methods such as \method{} that learn from natural language to work well in, and potentially address a larger gap in, less verifiable domains as well (such as personalization), though for the purposes of controlled study, we stick to verifiable ones.

\textbf{Retail.} We first introduce the Retail domain, a synthetic task carefully constructed to be verifiable and explicitly require the natural language context to solve any instance of the task. This domain further differs from prior studies on parametric learning in requiring the model to reason over the compositionality of its context and selectively apply relevant rules at inference time, rather than simply recall facts.

The model is tasked with calculating the price of a shopping cart given 30 discount rules that conditionally may apply. These rules are not provided at evaluation time when testing parametric learning methods. We programatically generate 256 evaluation queries as well as the corresponding ground truth price. Accuracy is measured as binary exact accuracy within $.01$ error. We provide a full example query and the entire list of natural language rules for the Retail domain in Appendix \ref{app:retail}.

\textbf{RuleArena (NBA).} We then turn to existing benchmarks. We take a task from the RuleArena benchmark ~\citep{zhou2025rulearenabenchmarkruleguidedreasoning}, typically intended as a complex instruction-following benchmark, and apply context distillation methods to internalize the entire corpus of natural language rules a model would normally be provided with in context. This corpus is approximately 20,000 tokens of NBA player trade rules and regulations to determine if a sequence of trades is illegal and if so why. We use the Level 2 setting for evaluation, which comprises the most difficult tasks in the benchmark. Accuracy is measured as exact match on the legality determination and violation identification.

\textbf{MTOB.} Though we focus our work on context that requires reasoning, for the sake of evaluating on very long context settings, we turn to the Machine Translation from One Book (MTOB) benchmark \citep{tanzer2024benchmarklearningtranslatenew}. Here, a model is tasked with translating an extremely low-resource language, Kalamang, for which there exists very limited resources outside of the corpus provided in the benchmark, to English. The grammar book and parallel examples span approximately 50,000 tokens, exceeding typical context windows and requiring 2$\times$ RoPE scaling for the ICL baseline. This domain emphasizes memorization more than compositional reasoning, making it a complementary challenge to Retail and RuleArena. For the MTOB domain specifically, we also omit the verification step. Because the model lacks prior knowledge of the target language, synthetic queries can only incorporate concepts present in the seed context. In contrast, for domains where the model has background knowledge, verification is necessary—the model might generate queries that implicitly require additional context beyond the seed. Performance is measured with the chrF metric \citep{popovic2015chrf}.

\subsection{\method{}: Scaling Data}
\label{sec:scaling_data}

\begin{figure*}[t!]
  \vskip -0.1in
  \begin{center}
  \centerline{\includegraphics[width=\linewidth]{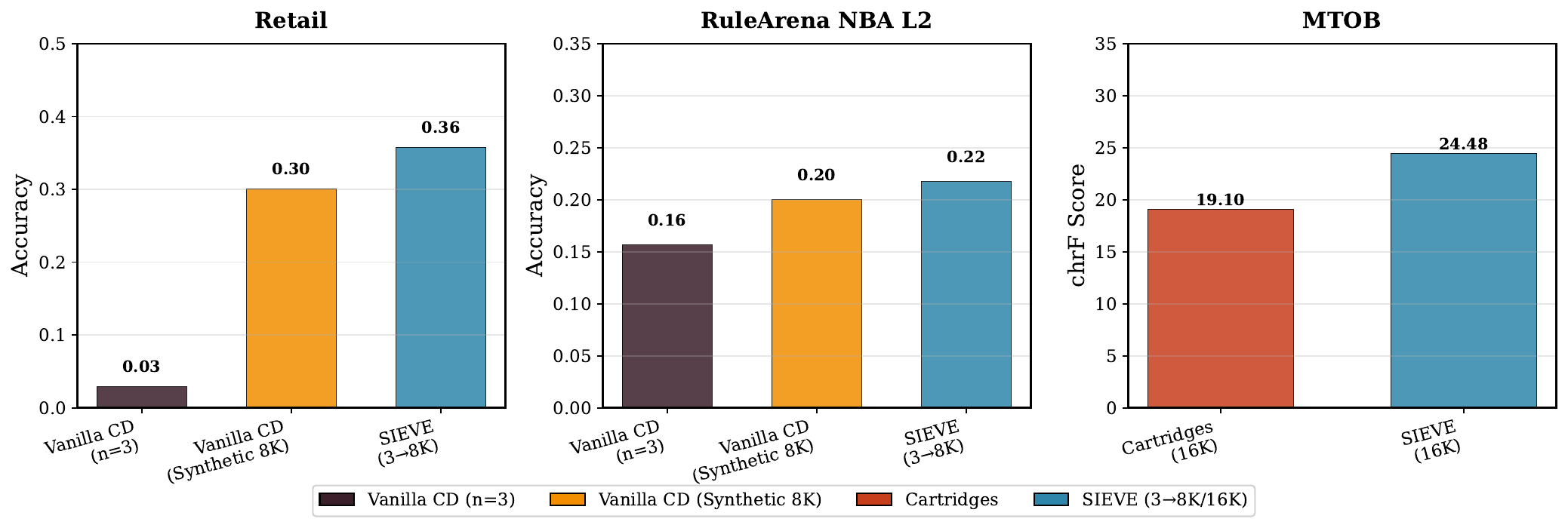}}
  \caption{\textbf{Comparison to baseline context distillation methods.} We compare \method{} against vanilla context distillation baselines across domains. $V_{CD}$ (3 seeds) trains on only the three seed query examples with all context. $V_{CD-S}$ (8K) uses our synthetically generated queries but includes all context during rollout generation (no selective filtering). \method{} generates synthetic data from three seeds to 8/16K scales and outperforms baselines with the same amount of training data in all scenarios.}
  \label{fig:baselines}
  \end{center}
  \vskip -0.25in
\end{figure*}

In \cref{fig:scaling_data}, we evaluate how \method{} scales with increasing amounts of synthetic data. Critically, all experiments across all domains use only the context and three query examples—no additional expert-curated traces or automated verifiers are required. These three examples mainly serve as formatting examples for the synthetic data generation. From the fixed context and three seeds, we evaluate \synthetic{} on up to 16,000 synthetic tuples of queries paired with their applicable context from the broader corpus of natural language context. We train for 2 epochs on the Retail and RuleArena domains and 5 epochs on the MTOB domain, chosen by empirically seeing when loss converged. Further details of training hyperparameters used by our runs of \method{} are provided in Appendix \ref{app:training_details}.

We compare our method with scale to an \textbf{in-context learning baseline} where the natural language context is provided directly in context at inference time. This is typically the gold baseline in traditional context distillation works, and what we aim to approximately match or potentially even surpass with our approach.

We consistently find across all three domains that performance improves as synthetic data scales, and trends suggest that scaling data generation further would continue to yield gains. Notably, \method{} roughly matches or outperforms ICL baselines at the scales we evaluate, with this effect most prominent in shorter-context settings (Retail). The performance further emphasizes that our method's data generation process allows for a trained model to compose different pieces of internalized context together at inference time, improving over ICL by 37.7\%.

For RuleArena and MTOB, the longer context nature (20-50K tokens requiring RoPE scaling) makes for more challenging tasks, yet \method{} still shows consistent improvement with scale and with 16K data points can match ICL performance without relying on any context.

These results demonstrate that sample-efficient parametric learning from natural language is achievable: with only three query examples, our method can internalize complex reasoning knowledge and match ICL performance without requiring context at inference time.

\subsection{Comparison to Baseline Context Distillation Methods} 
\label{sec:baselines}

We evaluate the effectiveness of \method{} in contrast to alternative context distillation approaches and show that it can outperform vanilla context distillation methods as well as methods designed for longer contexts. Results are shown in \cref{fig:baselines}.

\subsubsection{Baselines:}

\textbf{Vanilla Context Distillation ($V_{CD}$)}: Traditional context distillation methods, such as proposed in \citet{snell2022learningdistillingcontext, bhargava2024promptbaking}, do not leverage synthetic data towards sample-efficient learning. Thus we provide a result if we were to train on the same three seed query examples our method receives (and all context). Since vanilla context distillation does not perform synthetic data generation, this baseline operates under severe data constraints and is meant to show the importance of sample-efficient approaches. We train till loss converges.

\textbf{Vanilla Context Distillation w/ Synthetic Queries ($V_{CD-S}$)}: To create a fairer comparison, we provide vanilla context distillation with access to our synthetically generated queries (up to 8K examples). However, unlike \method{}, this baseline includes \emph{all} context in context during rollout generation, without filtering to only applicable context. As a result, this also serves as an ablation to the filtering portion of our method.

\textbf{Cartridges}: For MTOB, the 50K token corpus exceeds the model's 32K context window, making standard context distillation infeasible. We instead compare against Cartridges ~\citep{eyuboglu2025cartridgeslightweightgeneralpurposelong}, a method explicitly designed for long-context parametric memorization into KV caches, matching the amount of data to our method and using their recommended training hyperparameters. We do not evaluate Cartridges on Retail or RuleArena, as it is designed for factual recall from long documents rather than compositional reasoning over rules and instructions. Preliminary experiments suggested improvements over no-context baselines were marginal.

\textbf{Results.} On Retail, $V_{CD}$ with only three seed examples achieves just 3\% accuracy—insufficient data for the model to learn 30 compositional discount rules. When provided with 8K synthetic queries ($V_{CD-S}$), performance improves dramatically to 30\%, but \method{} still outperforms at 36\% by filtering to only applicable context during training. This demonstrates that \textbf{selective applicability is critical}: even with abundant synthetic queries, including all context indiscriminately yields worse results than our targeted approach.

On RuleArena (NBA), we observe similar trends: \method{} achieves a 10\% lift over $V_{CD-S}$ that uses our synthetic data and 37.5\% over a context distillation approach that only uses the three seeds.

For MTOB, vanilla context distillation is infeasible due to the 50K token corpus exceeding context limits. Instead, Cartridges achieves 19.10 chrF score—outperforming having no internalized knowledge (15.88)—while \method{} reaches 24.48 (both using 16K data points, though Cartridges targets the KV cache while \method{} targets the model weights). This represents a substantial improvement, though both methods fall short of the ICL baseline due to the difficulty of the long-context memorization task.

\textbf{Oracle Query Experiment.} To further isolate the importance of selective context filtering, we conduct an additional experiment on Retail where we programmatically generate queries from our ground truth pipeline rather than using synthetically generated queries — thus perfectly matching the train and test distributions. We provide these oracle queries to vanilla context distillation, which includes all context in context during rollout generation. Results in \cref{tab:oracle_retail} show that even with perfect queries, vanilla context distillation achieves only 27.11\%, while \method{} with synthetic queries achieves 33.98\%. This result underscores that selective applicability—pairing queries with only their relevant context—is potentially more important than query quality alone.

\begin{table}[t]
\caption{\textbf{Oracle access to programmatic queries on Retail.} Comparison under perfect access to programmatically generated queries, where the baseline method is trained on these ground truth (non-synthetic) queries and full context in the rollouts. Our selective context distillation outperforms vanilla context distillation even in this unfavorable setting where we use synthetic data. Both methods use 4096 examples for training.}
\label{tab:oracle_retail}
\vskip 0.15in
\begin{center}
\begin{small}
\begin{sc}
\begin{tabular}{lc}
\toprule
Method & Mean Accuracy (\%) \\
\midrule
Vanilla CD (Oracle Queries) & 27.11 \\
\method{} & \textbf{33.98} \\
\bottomrule
\end{tabular}
\end{sc}
\end{small}
\end{center}
\vskip -0.1in
\end{table}

\subsection{Ablations}
\label{sec:ablations}

\subsubsection{Multiple Rollouts vs. Scaling Data}
\label{sec:multiple_rollouts}

Recent work has suggested that generating multiple rollouts per query can improve distillation performance without requiring additional distinct data \citep{guha2025openthoughtsdatarecipesreasoning}. To investigate this tradeoff in our setting, we compare increasing the number of rollouts per query to scaling the number of distinct queries, while holding the total number of teacher generations approximately constant. Specifically, we compare two controlled settings: (i) generating 8 rollouts per query at smaller scales (512$\times$8 and 1024$\times$8), and (ii) scaling the number of distinct queries to matched totals (4096$\times$1 and 8192$\times$1). Results are shown in Table~\ref{tab:rollouts_vs_scale}.

We observe a tradeoff between scaling the number of distinct queries and increasing the number of rollouts per query. At smaller scales, increasing query diversity is more effective: 4096$\times$1 outperforms 512$\times$8, indicating that additional distinct queries provide greater benefit than repeated sampling of the same inputs. At larger scales, however, 1024$\times$8 slightly outperforms 8192$\times$1, suggesting that once sufficient query diversity has been achieved, multiple rollouts can provide complementary gains. This indicates a regime in which stochastic sampling of high-quality queries becomes increasingly beneficial.

\begin{table}[t]
\caption{\textbf{Multiple rollouts vs. scaling distinct queries on Retail.} We find that there is a tradeoff between generating multiple rollouts per query versus scaling the number of distinct queries, holding total generations constant, potentially indicating another axis for putting more compute in towards improved performance when data diversity is saturated.}
\label{tab:rollouts_vs_scale}
\vskip 0.15in
\begin{center}
\begin{small}
\begin{sc}
\begin{tabular}{lcc}
\toprule
Setting & Distinct Queries & Accuracy (\%) \\
\midrule
512$\times$8 & 512 & 30.23 \\
4096$\times$1 & 4096 & \textbf{33.98} \\
\midrule
1024$\times$8 & 1024 & \textbf{37.97} \\
8192$\times$1 & 8192 & 35.78 \\
\bottomrule
\end{tabular}
\end{sc}
\end{small}
\end{center}
\vskip -0.1in
\end{table}

Overall, these results suggest that scaling distinct queries is critical in low-data regimes, while multiple rollouts become a useful secondary mechanism for improved performance after query diversity saturates. The optimal tradeoff between query diversity and rollout count therefore appears to be domain- and scale-dependent, and should be determined empirically.

\subsubsection{Alternative Model Families}
\label{sec:model_families}

\begin{figure}[ht!]
  \centering
  \includegraphics[width=\columnwidth]{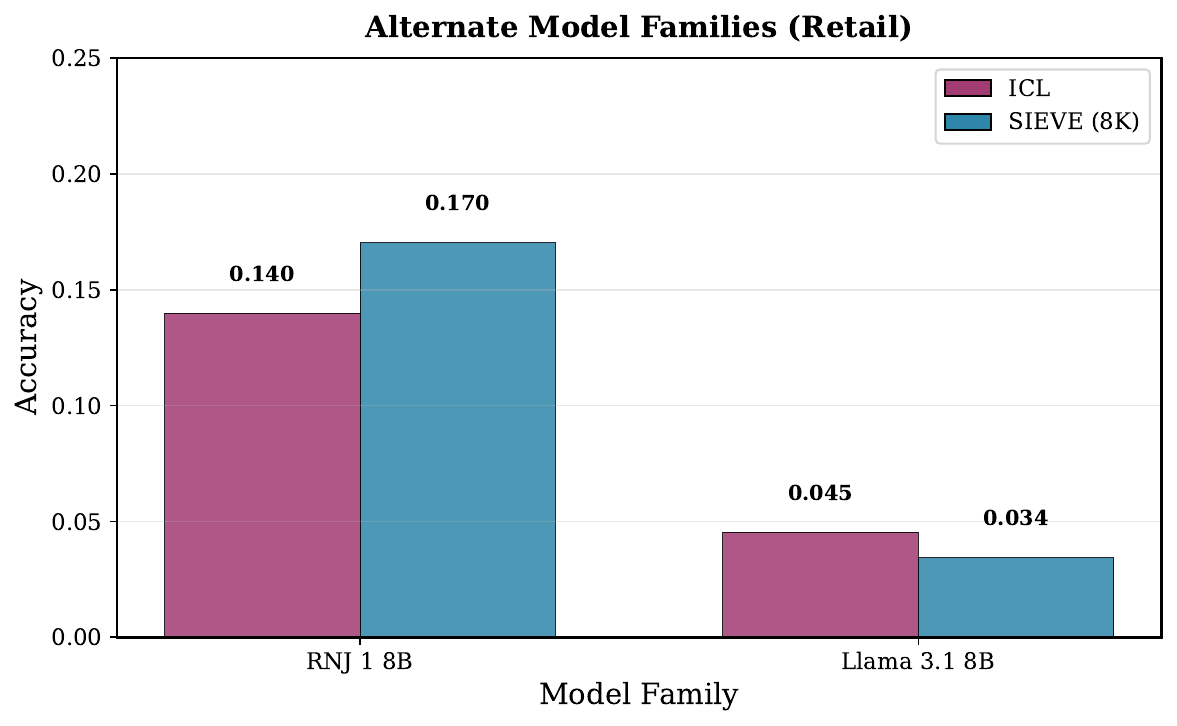}
  \caption{\textbf{\method{} generalizes across model families.} We evaluate \method{} on the Retail domain using alternative model families: Llama 3.1 8B and Rnj 1 8B. Results demonstrate that \method{} consistently improves model performance across diverse architectures (8K training examples).}
  \label{fig:model_families}
  \vskip -0.1in
\end{figure}

To verify that our method generalizes beyond the Qwen3 model family, we replicate experiments on the Retail domain using Llama 3.1 8B ~\citep{llama3} and Rnj 1 8B ~\citep{rnj1_base, rnj1_instruct}. Results are shown in \cref{fig:model_families}.

For Rnj 1 8B, \method{} achieves 17.03\% accuracy with 8K training examples, exceeding the ICL baseline of 13.98\%. This demonstrates that sample-efficient parametric learning from natural language context successfully generalizes to alternative model architectures, with the trained model outperforming ICL without requiring context at inference.

However, for Llama 3.1 8B, we observe a different pattern. The ICL baseline achieves only 4.53\% accuracy, and \method{} training yields 3.44\%—slightly below the ICL baseline. This reveals an important limitation: \textit{the base model must have sufficient capability for the method to be effective}. The Llama 3.1 8B model performs extremely poorly on Retail even with context in context, suggesting it lacks the foundational reasoning abilities required for this compositional rule-application task. When the base model is too weak, it struggles both to generate high-quality synthetic data through \synthetic{} and to internalize the resulting training signal effectively.

This ablation highlights that \method{} requires a model with reasonable capabilities on the target domain. For models that can leverage context in context (like Qwen3 and Rnj), our method successfully internalizes that context into weights. For models that cannot effectively use context even when provided (like Llama 3.1 8B on this task), sample-efficient parametric learning remains challenging. 

Future work could explore whether curriculum learning or progressive training strategies might enable weaker models to benefit from this approach. Another line of work might also consider decoupling the synthetic data generation models from the model that is trained, stepping away from the self-distillation paradigm we focus on in this work.
\section{Conclusion \& Future Work}
\label{sec:conclusion}

Our work introduces a method that achieves sample-efficient parametric learning from natural language. Through \method{} and its novel synthetic data generation method \synthetic{}, we show that models can internalize complex reasoning knowledge using a very small number of seed query examples. 
Our method outperforms prior context distillation methods in sample-efficient settings. 
\method{} further matches or exceeds the performance of in-context learning without requiring context at inference time. Our key insight is that for many tasks, not all context applies to every query. 
By decomposing the context and constructing synthetic data that only use applicable parts of it, we sample higher quality rollouts that outperform past context distillation methods, including when given access to more data or higher-quality (non-synthetic) oracle queries.

We find that parametric learning can be practical across multiple domains requiring compositional reasoning (Retail, RuleArena) and long-context memorization (MTOB).  
Crucially, this can be achieved without the complexities that traditionally make parametric learning challenging: many examples of input queries, large supervised datasets of expert-curated traces, or verifiers. This opens new possibilities for continual learning systems that improve persistently from natural language context in real-world settings.

\textbf{Future Directions.} Sample-efficient parametric learning from natural language enables several promising research directions. First, this paradigm allows for learning algorithms that adapt from rich, non-scalar context in sample-efficient ways moving beyond simple preference scores to leverage the full expressiveness of natural language. Second, expanding the suite of learning methods remains important: exploring architectural modifications (e.g., specialized memory layers), alternative training objectives beyond context distillation, and methods for managing interference between updates could further improve sample efficiency and scalability. Third, developing more challenging benchmarks that require deeper reasoning, longer-term memory, and compositional generalization will be critical for advancing parametric learning methods. Finally, studying how these sample-efficient learning methods can enable truly continual learning—where models iteratively improve from ongoing feedback—represents an exciting frontier for making LLMs more adaptive and specialized over time.
\section*{Acknowledgements}

We would like to thank Alan Zhu, Andrew Qin, and Joseph E. Gonzalez for many insightful discussions about this work. 

Sky Computing Lab is supported by gifts from Accenture, Amazon, AMD, Anyscale, Broadcom, Google, IBM, Intel, Intesa Sanpaolo, Lambda, Lightspeed, Mibura, NVIDIA, Samsung SDS, and SAP. We would additionally like to acknowledge Databricks for their generous compute support.

This material is based upon work supported by the National Science Foundation Grant No. CCF-2019844. Any opinions, findings, and conclusions or recommendations expressed in this material are those of the author(s) and do not necessarily reflect the views of the National Science Foundation.

% \section*{Impact Statement}

% This paper presents work whose goal is to advance the field of Machine
% Learning. There are many potential societal consequences of our work, none
% which we feel must be specifically highlighted here.

\bibliography{main}
\bibliographystyle{icml2026}

%%%%%%%%%%%%%%%%%%%%%%%%%%%%%%%%%%%%%%%%%%%%%%%%%%%%%%%%%%%%%%%%%%%%%%%%%%%%%%%
%%%%%%%%%%%%%%%%%%%%%%%%%%%%%%%%%%%%%%%%%%%%%%%%%%%%%%%%%%%%%%%%%%%%%%%%%%%%%%%
% APPENDIX
%%%%%%%%%%%%%%%%%%%%%%%%%%%%%%%%%%%%%%%%%%%%%%%%%%%%%%%%%%%%%%%%%%%%%%%%%%%%%%%
%%%%%%%%%%%%%%%%%%%%%%%%%%%%%%%%%%%%%%%%%%%%%%%%%%%%%%%%%%%%%%%%%%%%%%%%%%%%%%%
\newpage
\appendix
\onecolumn
\section{Training Details}
\label{app:training_details}

We run all training experiments unless otherwise noted on a single node of 8xH100s. For reproducibility, we provide key non-default training parameters used in \cref{tab:soft_distillation_hyperparams} and \cref{tab:soft_distillation_epochs}. Number of epochs were chosen empirically per domain based on loss convergence.

\begin{table}[h]
\centering
\caption{Training Hyperparameters}
\label{tab:soft_distillation_hyperparams}
\begin{tabular}{lc}
\toprule
\textbf{Hyperparameter} & \textbf{Value} \\
\midrule
Learning Rate & $1 \times 10^{-5}$ \\
Batch Size (per device) & 1 \\
Gradient Accumulation & 8 \\
Effective Batch Size & 64 \\
Temperature ($\tau$) & 1.0 \\
Warmup Steps & 50 \\
Top-K Tokens & 100 \\
Max Sequence Length & 16{,}384 \\
Optimizer & AdamW \\
DeepSpeed Config & ZeRO-3 \\
Number of GPUs & 8 \\
\bottomrule
\end{tabular}
\end{table}

\begin{table}[h]
\centering
\caption{Dataset-Specific Hyperparameters}
\label{tab:soft_distillation_epochs}
\begin{tabular}{lccc}
\toprule
\textbf{Hyperparameter} & \textbf{Retail} & \textbf{NBA} & \textbf{MTOB} \\
\midrule
Epochs & 2 & 2 & 5 \\
\bottomrule
\end{tabular}
\end{table}
\section{Extended Domain Discussion}
\label{app:domain_discussion}

We provide further information regarding the domains we evaluate on in the paper here.

\subsection{Retail}
\label{app:retail}

\textbf{Retail Domain.} The Retail domain involves calculating final prices after applying complex discount rules to shopping carts. Each query is programmatically generated with a subset of the following components:

\begin{itemize}
    \item \textbf{Customer Types:} 6 types (student, senior citizen, veteran, employee, teacher, regular)
    \item \textbf{Product Categories:} 8 categories (electronics, clothing, books, food, home, sports, beauty, health)
    \item \textbf{Promo Codes:} 7 promotional codes (SAVE20, WELCOME10, STUDENT15, HOLIDAY30, NEWBIE5, TEACHER10, BULK10)
    \item \textbf{Membership Tiers:} 3 tiers (bronze: 1+ years, silver: 3+ years, gold: 5+ years)
\end{itemize}

Each query presents a customer profile (type, membership years), shopping cart (1-5 items with categories, prices, quantities), and optional promo code. The model must calculate the final price after applying all applicable discount rules, which may stack multiplicatively or apply to specific categories. Ground truth is computed programmatically using the rule engine.

\begin{tcolorbox}[colback=blue!5!white,colframe=blue!60!white,title=Retail Example Query]
Calculate the final price for the following customer purchase after applying all applicable discount rules.\\

Customer Profile:\\
- Type: senior\\
- Membership years: 4\\

Shopping Cart:\\
- Shoes (apparel): \$85.00 x 2\\
- Jacket (apparel): \$60.00 x 1\\
- Coffee Maker (home): \$45.00 x 1\\

Promo code: None\\

IMPORTANT: Apply discounts in this exact order to the running total:\\
1. Category-specific percentage discounts (apply only the highest discount per category to each category's subtotal)\\
2. Total purchase percentage discounts (apply only the highest total discount to the remaining amount after step 1)\\
3. Fixed amount discounts (subtract from the remaining amount after step 2, sum all applicable fixed discounts)\\

Note: Each discount applies to the current running total, not the original price.
\end{tcolorbox}

\begin{tcolorbox}[colback=blue!5!white,colframe=blue!60!white,title=Retail Rules]
Discount Rules:\\
- If customer is a student AND total spend is at least \$50, apply 10\% discount to total purchase\\
- If customer is a senior citizen AND total spend is at least \$50, apply 15\% discount to total purchase\\
- If customer is a employee AND total spend is at least \$50, apply 20\% discount to total purchase\\
- If customer is a teacher AND total spend is at least \$50, apply 10\% discount to total purchase\\
- If customer is a student AND cart contains electronics, apply 15\% discount on electronics items only\\
- If customer is a student AND cart contains books, apply 15\% discount on books items only\\
- If customer is a senior citizen AND cart contains food, apply 20\% discount on food items only\\
- If customer is a veteran AND cart contains electronics, apply 5\% discount on electronics items only\\
- If customer is a employee AND cart contains home, apply 25\% discount on home items only\\
- If customer is a teacher AND cart contains books, apply 15\% discount on books items only\\
- If promo code is 'SAVE20' AND total spend is at least \$100, apply 20\% discount to total purchase\\
- If promo code is 'WELCOME10', apply \$10 fixed discount\\
- If customer is a student AND promo code is 'STUDENT15', apply 15\% discount to total purchase\\
- If promo code is 'HOLIDAY30' AND total spend is at least \$150, apply 30\% discount to total purchase\\
- If promo code is 'NEWBIE5', apply \$5 fixed discount\\
- If customer is a teacher AND promo code is 'TEACHER10', apply 10\% discount to total purchase\\
- If promo code is 'BULK10' AND total spend is at least \$200, apply 10\% discount to total purchase\\
- If total spend is at least \$150, apply 5\% discount to total purchase\\
- If total spend is at least \$200, apply 10\% discount to total purchase\\
- If cart contains electronics items AND total electronics spend is \$300 or greater, apply 10\% discount on electronics items only\\
- If cart contains clothing items AND total clothing spend is \$75 or greater, apply 10\% discount on clothing items only\\
- If cart contains books items AND total books spend is \$45 or greater, apply 10\% discount on books items only\\
- If cart contains food items AND total food spend is \$30 or greater, apply 10\% discount on food items only\\
- If cart contains home items AND total home spend is \$60 or greater, apply 10\% discount on home items only\\
- If cart contains sports items AND total sports spend is \$90 or greater, apply 10\% discount on sports items only\\
- If cart contains beauty items AND total beauty spend is \$52 or greater, apply 10\% discount on beauty items only\\
- If cart contains health items AND total health spend is \$75 or greater, apply 10\% discount on health items only\\
- If customer has been a member for 1 or more years, apply 5\% discount to total purchase\\
- If customer has been a member for 3 or more years, apply 10\% discount to total purchase\\
- If customer has been a member for 5 or more years, apply 15\% discount to total purchase\\
\end{tcolorbox}

Some of the example context units that the natural language context may decompose into for this domain include:

\lstset{
  basicstyle=\ttfamily\small,
  breaklines=true,
  breakatwhitespace=true,
  frame=single
}

\begin{lstlisting}
["If customer is a student AND total spend is at least $50, apply 10% discount to total purchase",
 "If customer is a senior citizen AND total spend is at least $50, apply 15% discount to total purchase",
 "If customer is an employee AND total spend is at least $50, apply 20% discount to total purchase",
 ...]
\end{lstlisting}

\clearpage

\subsection{RuleArena (NBA)}
\label{app:nba}

Quoting the original description of this domain from the benchmark: "It requires LLMs to determine whether one or more specified transactions are allowed. The regulations are extracted from the 2023 NBA Collective Bargaining Agreements (CBA) and excerpt from the NBA Constitution and ByLaws. Complexity arises from the numerous factors influencing transaction eligibility, including the player’s contract value, salary-matching constraints, and the specific transaction date. LLMs must accurately identify and apply the relevant rules from the agreement to determine whether a given transaction can proceed" ~\citep{zhou2025rulearenabenchmarkruleguidedreasoning}.

\begin{tcolorbox}[colback=blue!5!white,colframe=blue!60!white,title=NBA Example Query]
QUESTION:\\
Team Situations:\\
Team A has a team salary of \$130,000,000.\\
Team B has a team salary of \$135,000,000.\\
Team C has a team salary of \$98,000,000.\\

Player Situations:
Player A was the 33th second-round pick of Team B in 2022 NBA draft when he was 20 years old.\\
Player A signed a 2-year contract with Team B providing the minimum salary.\\
Player B was the 44th second-round pick of Team C in 2015 NBA draft when he was 22 years old.\\
Player B signed a 3-year contract with Team B providing annual salary \$31,000,000, 5\% increase per year in 2021 Cap Year.\\
Player C was the 21th first-round pick of Team C in 2016 NBA draft when he was 22 years old.\\
Player C signed a 2-year contract with Team B providing annual salary \$7,000,000, 5\% increase per year in 2022 Cap Year.\\
Player D was the 15th first-round pick of Team D in 2017 NBA draft when he was 22 years old.\\
Player D signed a 3-year contract with Team D providing annual salary \$10,000,000, 5\% increase per year in 2021 Cap Year.\\

Operations:
A. Team B provides a qualifying offer for Player A.\\
B. Team A provides Player A with an offer sheet - a 2-year contract providing annual salary \$10,000,000 in the first Salary Cap Year (2024-2025), 5\% increase per year for the first two Salary Cap Years.\\
C. Team B matches the offer by team A.\\
D. Team B signs a 2-year contract with Player D providing annual salary \$5,000,000 in the first Salary Cap Year (2024-2025), 5\% increase per year.\\
E. Team C signs a 2-year contract with Player C providing annual salary \$7,000,000 in the first Salary Cap Year (2024-2025), 5\% increase per year.\\
\end{tcolorbox}

The full set of rules for this domain is too large to include in the paper, thus we direct readers to the official repository here: \url{https://github.com/SkyRiver-2000/RuleArena/blob/main/nba/reference_rules.txt}.

Some of the example context units that the natural language context may decompose into for this domain include:

\lstset{
  basicstyle=\ttfamily\small,
  breaklines=true,
  breakatwhitespace=true,
  frame=single
}

\begin{lstlisting}
["the Team shall be prohibited from trading (either conditionally or unconditionally) its first round draft pick in the first NBA Draft that occurs following the seventh Season that follows the Season occurring within such Salary Cap Year;", "A Team may use the Taxpayer Mid-Level Salary Exception to sign one (1) or more Player Contracts during each Salary Cap Year not to exceed two (2) Seasons in length, that, in the aggregate, provide for Salaries and Unlikely Bonuses in the first Salary Cap Year totaling up to the amounts set forth below, provided that the Team's Team Salary immediately following the Team's use of such Exception exceeds the First Apron Level:", "Player Contracts signed pursuant to the Mid-Level Salary Exception for Room Teams may provide for annual increases and decreases in Salary and Unlikely Bonuses in accordance with Section 5(a)(1) above.", ...]
\end{lstlisting}

\clearpage

\subsection{MTOB}
\label{app:mtob}

The Machine Translation from One Book (MTOB) benchmark ~\citep{tanzer2024benchmarklearningtranslatenew} tests a model's ability to translate between English and Kalamang, an extremely low-resource language for which very few resources exist on the internet (essentially a single grammar book) that they curate into the data provided in this benchmark.

We make use of their medium sized version of this grammar book for our natural language context in this task, which comprises of approximately 50K tokens that include a grammar table and numerous word mappings. We also include a corpus of approximately 375 paired English-Kalamang sentences. We use RoPE scaling to extend the model's context window for ICL baseline evaluations.

We refrain from including example queries and decompositions of the natural language corpus here as it is both too long but more importantly not recommended to prevent leakage in the benchmark. At a high level though, in this setting the context units include both groups of word translations, individual grammar rules, and example sentences. 
\section{Synthetic Data Prompts}
\label{app:synthetic_data_prompts}

Below we provide the general prompts used for all domains in \method{} (that fit within context).

\subsection{Decomposition Prompt}
\begin{tcolorbox}[colback=red!5!white,
                  colframe=red!60!white,title=Decomposition Prompt]

Break down the following feedback/guidelines/knowledge into atomic, independent items.\\

Each atomic item should:\\
1. Express a single, self-contained rule, fact, definition, or example\\
2. Be evaluable independently (can determine if it applies without needing other items)\\
3. Preserve the exact meaning and wording from the original\\

Content:\\
\{chunk\}\\

Output each atomic item separated by "\#\#\#" on its own line.\\
For items with sub-bullets or multiple lines, include all lines as part of that item.
Do NOT number or label items. Do not add explanations or commentary.\\

Example format:\\
First item content here\\
\#\#\#\\
Second item content here\\
\#\#\#\\
Third item content here\\

Do not group multiple concepts together. Each item should be atomic.\\

\end{tcolorbox}

\subsection{Seed Context Selection Prompt (Base Model)}
\begin{tcolorbox}[colback=red!5!white,
                  colframe=red!60!white,title=Seed Context Selection Prompt]

Task: Select 3-5 guidelines from the natural language feedback that could apply together to a single scenario/question.\\

Guidelines:\\
\{feedback\}\\
\{examples section\}\\
Selected guidelines:\\
-\\

\end{tcolorbox}

\subsection{Query Generation Prompt}
\begin{tcolorbox}[colback=red!5!white,
                  colframe=red!60!white,title=Query Generation Prompt]

Generate a realistic question where the following feedback/guidelines/knowledge would apply:\\

\{selected feedback\}\\

Instructions:\\
1. Create a specific question where the information applies, similar to the format of the examples below\\
2. Make it realistic\\
3. Include all necessary details\\
4. Output ONLY the question, nothing else\\

\{examples section\}\\

Question:\\

\end{tcolorbox}
%%%%%%%%%%%%%%%%%%%%%%%%%%%%%%%%%%%%%%%%%%%%%%%%%%%%%%%%%%%%

%%%%%%%%%%%%%%%%%%%%%%%%%%%%%%%%%%%%%%%%%%%%%%%%%%%%%%%%%%%%%%%%%%%%%%%%%%%%%%%
%%%%%%%%%%%%%%%%%%%%%%%%%%%%%%%%%%%%%%%%%%%%%%%%%%%%%%%%%%%%%%%%%%%%%%%%%%%%%%%

\end{document}